\documentclass[10pt,letterpaper]{article}

\usepackage{cogsci}

\cogscifinalcopy 

\usepackage{pslatex}
\usepackage{apacite}
\usepackage{float} 

\usepackage{algorithm}
\usepackage{algorithmic}
\usepackage{multirow}
\usepackage{booktabs}
\usepackage{enumitem}
\usepackage{epigraph}
\usepackage{csquotes}
\usepackage[dvipsnames]{xcolor}
\usepackage{tcolorbox}
\usepackage{amsmath}
\usepackage{natbib}
\usepackage[symbol]{footmisc}

\title{Chain of Thought Still Thinks Fast: APriCoT Helps with Thinking Slow}

\author{
  {\large \bf Kyle Moore}$^\dagger$ \\
  Vanderbilt University \\
  kyle.a.moore@vanderbilt.edu
  \And
  {\large \bf Jesse Roberts}$^\dagger$ \\
  Tennessee Technological University \\
  Vanderbilt University \\
  jtroberts@tntech.edu
  \AND
  {\large \bf Thao Pham} \\
  Berea College
  \And
  {\large \bf Douglas Fisher} \\
  Vanderbilt University
  }

\begin{document}

\maketitle

\begin{abstract}
Language models are known to absorb biases from their training data, leading to predictions driven by statistical regularities rather than semantic relevance. We investigate the impact of these biases on answer choice preferences in the Massive Multi-Task Language Understanding (MMLU) task. Our findings show that these biases are predictive of model preference and mirror human test-taking strategies even when chain of thought (CoT) reasoning is used. To address this issue, we introduce Counterfactual Prompting with Agnostically Primed CoT (APriCoT). We demonstrate that while Counterfactual Prompting with CoT alone is insufficient to mitigate bias, APriCoT effectively reduces the influence of base-rate probabilities while improving overall accuracy. Our results suggest that mitigating bias requires a slow thinking process which CoT alone may not provide as it tends to reinforce fast thinking model bias under some prompting methodologies. APriCoT is a step toward developing more robust and fair language models that can \textit{think slow}.


\end{abstract}

\section{Introduction}

A\footnote[0]{$\dagger$ equal contribution} model of human cognitive reasoning involving two distinct modes of thinking was posited in \citeauthor{kahneman2011thinking}'s seminal work, \textit{Thinking Fast and Slow} \citep{kahneman2011thinking}. Thinking fast, or heuristically, is performed by System 1, while thinking slow, or deliberatively, is performed by System 2. Although this dichotomy has received both support and criticism since its publication, the concepts of System 1 and System 2 thought remain useful paradigms, evoking the relationship between single-shot response and recursive generation.

We examine base rate probability (BRP) effects closely in the LLaMa 3.1 8B model \citep{dubey2024llama} and find a consistent BRP bias, consistent with existing findings for other models. Past work has shown this leads to responses that are based on bias apart from any reasoning  \citep{moore2024base}. We attempt to mitigate bias driven response and encourage reasoning by combining counterfactual (CF) prompting with chain of thought (CoT) \citep{wei2022chain}. Surprisingly, this exacerbates the issue.




To understand why this confirmation bias-like behavior occurs, we formulate Agnostically Primed Chain of Thought (APriCoT) based on the theoretical explanation of CoT provided by \citet{prystawski2024think}. Their work suggests that CoT provides a method of traversing the semantic space but is not untrustworthy when the question provides insufficient priming  \citep{turpin2024language}. To address this, APriCoT forces the model to reason down multiple semantic directions.

Evaluating model preference over all resultant semantic paths is taken as an estimate of the model's preferred response. We find this response is distributed nearly identically to the ground truth answer distribution and has better accuracy than either CF or CoT, showing APriCoT helps the model to think more \textit{slowly}.


In this paper we make 3 primary contributions: 1) We extend CF prompting to accommodate CoT reasoning, enabling a more comprehensive analysis of BRP effects. 2) We demonstrate that CoT reasoning, when adopting CF prompting strategies, exacerbates BRP effects on overt behavior, challenging the widely accepted utility of CoT. 3) We introduce an extension to CoT prompting called APriCoT which reduces BRP effects and improves accuracy. We posit that APriCoT may serve as an effective strategy for evaluating model behaviors and performing complex reasoning tasks.



    





\section{Prior Work}

In this section, we briefly survey the existing literature surround mitigation of base rate probability (BRP) effects and the use of CoT in connection with other heuristics and biases. BRP effects are defined and described in detail in a dedicated section shortly, but can briefly be understood as effects of word preferences on overt behavior while performing unrelated tasks. BRP effects and methods for mitigating them in non-CoT contexts were evaluated in some depth by \citet{moore2024base}, \citet{wei2024unveiling}, and \citet{zheng2023large}. Both works found that LLMs are prone to BRP effects in the context of MCQA tasks. \citet{moore2024base} proposed a method for reformulating MCQA tasks to be less sensitive in to BRP effects. \citet{wei2024unveiling, zheng2023large} attempted to mitigate the BRP effect in behavior by calibrating the output probabilities using the measured BRPs. \citet{chua2024bias} attempted to mitigate LLM initial behavior biases in a CoT context using fine-tuning, but the biases being mitigated were user-induced rather than inherent to the model.

Prior research has also identified similar undesirable behaviors with CoT. \citet{shaikh2023second} found that CoT is prone to reveal and exacerbate the social biases that the model learned during pre-training. This hints at a consistent bias reinforcement behavior across bias dimensions. \citet{saparov2023language} investigated CoT's utility for complex reasoning and found that models using CoT are capable of strong instantaneous deductive steps but rarely explore multiple semantic paths for more deliberative reasoning, which APriCoT attempts to induce explicitly. 

Finally, this paper extends the significant body of work studying LLM cognition. Prior work has investigated fan effects \citep{roberts2024large2}, typicality effects \citep{roberts2024using, misra2021language}, and theory of mind \citep{ullman2023large}.

\section{Background}

This section provides the pedagogical references and supporting literature on which the rest of the paper depends.

\textbf{The Massive Multitask Language Understanding} (MMLU) task \citep{hendrycks2020measuring} is a MCQA benchmark composed of 14042 questions distributed unevenly across 57 subjects. Each question has four answer choices provided, with one answer choice designated as correct. MMLU is a standard benchmark included in many model release publications and is intended to quantitatively measure a model's natural language understanding and factual knowledge. The ubiquity of MMLU highlights the importance of robust testing methods that allow evaluators to isolate measured behaviors from entangled variables like the BRP effect.

MMLU places few restrictions on the prompting method used for evaluation, making it deceptively difficult to compare accuracy measures across models. Evaluation methods often include optional in-context learning up to 5-shots and CoT reasoning. Questions used for 5-shot in-context learning examples are provided as a separate set and are not included in the reported benchmark accuracies.

\textbf{Cloze testing} is a prompting strategy commonly used to evaluate tasks where LLMs must select from a set of options. In this approach, the probability of each option is measured given a shared context. The chosen option is taken to be the option that yields the highest probability, i.e. $max_{a \in A}P(a | C)$ where $C$ is the task context and $A$ is the set of available option signifiers. For example, a four option multiple choice question answering (MCQA) task can be evaluated via cloze testing by providing the question and answer choices to the model before finishing with a query statement similar to ``The most correct answer choice is". The question, choices, and query are provided to the model and the probabilities of the four completions ``A", ``B", ``C", and ``D'' are computed independently. The model's answer is determined by the option with the highest probability.

\textbf{Counterfactual (CF) prompting} differs in the structure of the prompt. Instead of a shared query, CF prompting uses a separate query for every option and a shared \textit{canary} completion for evaluation. The option eliciting the highest canary probability is taken as the model's choice, i.e. $max_{a \in A}P(canary | C, a)$. For a multiple choice question, the question and answer choices remain unchanged, while the query portion of the prompt takes a form similar to ``Answer choice X is the most", where X is one of ``A", ``B", ``C", or ``D", and the canary completion for all options is ``correct".

CF prompting is intended to mitigate BRP effects while maintaining task semantics. Prior work has shown that CF evaluation fails to strongly mitigate BRP effects and may degrade performance on some tasks \citep{moore2024base}. Because the task definition semantic content is equivalent between CF and cloze prompting, expected model behavior measurements should not change. 

Change in task performance with equivalent prompting methods is indicative of fragile, under-supported behaviors in the model. For the MCQA example, significant loss in performance from simple rewording of the questions, rearranging of answer choices, or changing the wording when asking the model for its answer suggests that the model does not truly possess either sufficient question understanding and/or the requisite factual knowledge.

\textbf{Chain-of-Thought (CoT)} is a common prompting strategy proposed by \citet{wei2022chain} in which the language model is encouraged to generate freely from a task definition before providing a final answer. It is intended to approximate System-2-like processing and has been shown to improve model performance on a wide array of difficult benchmarks across numerous tasks \citep{suzgun2023challenging}.

The mechanism by which CoT yields improvement is not yet fully understood and subject to ongoing research. \citet{madaan2023makes} suggested that it functions, in part, by improving a model's understanding of the target task by retrieving and incorporating relevant information into the context. This seems to be in line with \citet{prystawski2024think}, who theorize that CoT works by traversal of a Bayes net like semantic space formed by the union of the pre-training data, the provided context, and a target completion. Models may not traverse the optimal path in this space and have been shown in \citet{saparov2023language} to be greedy reasoners that struggle to explore multiple pathways and instead tend to traverse toward pre-existing or induced biases \citep{turpin2024language}.

\subsection{Base-Rate Probability Effects}
LLMs predict the most likely next token given all tokens in the preceding context. This conceptually simple function has shown since at least \citet{radford2019language} to be sufficient to perform a wide variety of tasks with a surprising level of competency. While tokens have an in-context probability, they also have an intrinsic probability called the base-rate probability (BRP) \citep{moore2024base}. 

BRP is the probability associated with a token given a minimal amount of context necessary to isolate the represented concept. Differences in BRP between tokens have been shown to affect models' overt behavior and performance on downstream tasks both when prompted via cloze and CF prompting \citep{moore2024base}. BRP is conceptually linked to the notion of priors in Bayesian models. Prior work has sought to measure Bayesian priors in LLMs and found that the priors exhibited by LLMs are qualitatively human-like across a variety of contexts \citep{zhu2024eliciting}.

BRP effects may make it difficult to isolate behaviors and abilities that are being measured. For example, a model may be evaluated as being more prone to violence if it is evaluated by a yes/no questionnaire in which violent-indicative yeses outnumber peaceful-indicative yeses simply because the BRP for yes is higher than that for no. Given the same task with a trivial reversal of the question polarity may yield reversed results without semantic changes to the context.

\begin{table}[h]
\centering
\begin{tabular}{l}
\textbf{BRP Prompt Template} \\
\toprule
\begin{tabular}[c]{@{}l@{}}
Below you will see a question and answer choices.\\
QUESTION\\
choice \textless{}1st label\textgreater\ CHOICE\\
choice \textless{}2nd label\textgreater\ CHOICE\\ 
choice \textless{}3rd label\textgreater\ CHOICE\\
choice \textless{}4th label\textgreater\ CHOICE\\
In conclusion, which choice do you believe\\
is most correct?\\
I believe the correct answer is choice \rule{0.5cm}{0.15mm}
\end{tabular}
\\ \bottomrule

\end{tabular}
\caption{Prompt template used for measuring BRP. 1st-4th labels are replaced iteratively with every ordering of the choices ``A", ``B", ``C", ``D". The probabilities of each answer choice is measured and averaged across all orderings.}
\label{tab:BRP-prompt}
\end{table}

\section{LLaMa-3.1 Token Base-Rate Probability}

\begin{table*}[t]
\centering
\begin{tabular}{l c c}
\textbf{Pre-CoT Prompt Template} & \textbf{Post-CoT Prompt Template} & \textbf{Canary Word} \\ \toprule

\begin{tabular}[c]{@{}l@{}}Below you will see a question and answer choices.\\
\textless{}question text\textgreater\\
choice A: \textless{}1st answer choice\textgreater\\
choice B: \textless{}2nd answer choice\textgreater\\ 
choice C: \textless{}3rd answer choice\textgreater\\
choice D: \textless{}4th answer choice\textgreater\\
\textcolor{red}{\framebox{CoT}\ \ \ Let's think step by step.}\\
\textcolor{ForestGreen}{\framebox{APriCoT}\ \ \ Let's evaluate choice \textless{}N-th choice\textgreater\ step by step.}
\end{tabular} &

\begin{tabular}[c]{@{}l@{}}
In conclusion, do you believe choice\\
\textless{}N-th choice\textgreater\ is most correct?

\end{tabular} &
Yes
\\ \bottomrule

\end{tabular}
\caption{Prompt templates used for each prompting method. \textcolor{red}{Red text} denotes context that is present only in CoT. \textcolor{ForestGreen}{Green text} denotes context that is present only in APriCoT.}
\label{tab:prompts}
\end{table*}
In this section, we describe how BRPs were obtained for the LLaMa 3.1 model. BRP can be naively defined as the probability of a token given no prior context, but may also be contextually defined to account for BRP differences between concepts that share the same token. For example, when measuring the BRP of the letter ``A", it is important to differentiate between the BRP of the article ``A'' and ``A'' as an answer choice in a MCQA task. This example is especially noteworthy, as the article ``A'' has a high contextual probability with an empty context string because ``A'' is among the most common words to start a sentence or even entire documents.

The remainder of this paper investigates prompting strategies using the common MMLU benchmark as a test-bed, meaning that we must evaluate the relative BRP of each answer choice in the context of an MCQA task. We use nearly the same method as described in \citet{moore2024base} to calculate BRPs, changing only the prompt template to match the templates described in later sections and summarized in Table \ref{tab:prompts}. We measure BRPs using cloze prompting in line with previous work, which found the cloze BRP was predictive of overt behavior for both cloze and CF. We verify this claim by measuring the correlation between cloze BRP and overt behavior later in this work. The template used to measure BRP is described in Table \ref{tab:BRP-prompt}. Our results suggest that the model has a BRP preference ordering of $A \succeq D \succ B \succeq C$.



In \citet{moore2024base}, the authors found that CF prompting weakly mitigates BRP effects on overt behavior, but fails to eliminate them entirely. We hypothesize this is because the models are restricted to immediately completing the task, with no ability to reason over the potential answers or question for more complex questions. This forces the models to resort to the heuristics like answer choice or positional preference. Heuristic behaviors of this type are reminiscent of System-1 processing \citep{kahneman2011thinking}. 

In line with the two system theory, we contend that a more deliberative, System-2-like, evaluation will eliminate the BRP effect. To evaluate this, we propose a novel adaptation of the CF methods used in \citet{moore2024base} to include chain-of-thought reasoning.

\subsection{Methods: CF+CoT}
We extend existing work by evaluating the LLaMa 3.1 8B model for its susceptibility to BRP effects under CF+CoT reasoning through the use of a common multiple choice question answering (MCQA) benchmark, MMLU.


For comparison purposes, we evaluate the model using both CoT with CF evaluation and CF evaluation with no CoT. We refer to these as experiments as CF+CoT and CF respectively for the rest of this work. In both cases, the model is presented with the question and answer choices using the second-person, conversational prompt template shown in Table \ref{tab:prompts}. In line with the methods used in \citet{wei2022chain}, we elicit CoT reasoning by using the sentence ``Let's think step by step.'' The model is then allowed to generate up to 100 tokens of CoT reasoning. In both cases, the model is queried for it's answer using CF prompting. To mitigate the effects of cutting off reasoning mid-sentence, we always begin the query, shown in the center column of Table \ref{tab:prompts}, with a new line character and the words "In conclusion".

Trials also include the clause ``In conclusion,'' to signal the end of the CoT section. This is also true for CF trials so as to isolate the effects of adding CoT. Four copies (one per answer choice) of this combined context are input to the model, differing only by the answer choice presented in the query. The probability of the canary token ``Yes'' is evaluated given the four contexts. To promote robustness in the results, we repeat this process 10 times per question, taking the mean probability for each answer choice over all 10 trials. The model's chosen answer choice is interpreted as the answer choice that yields the highest mean probability across all trials.


\subsection{Results: CF+CoT}

\begin{figure}[H]
    \centering
    \includegraphics[width=0.9\linewidth]{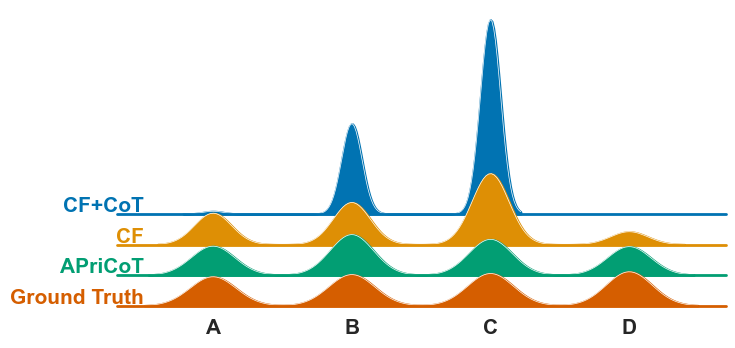}
    \caption{Chosen MMLU answer choice distribution for each prompting method and ground truth. CF+CoT tends to skew the distribution further from the ground truth. APriCoT tends to flatten the distribution toward ground truth.}
    \label{fig:dist-ridge}
\end{figure}

We measured the answer choice distribution in the responses from the model using both prompting strategies across all choice orderings. The results of this are shown in the first two rows of Figure \ref{fig:dist-ridge}. Consistent with previous results in \citet{moore2024base}, we see a strong preference for one answer choice over all others in the case of CF reasoning. In general, the model, when evaluated without CoT, shows a strong preference for choice C and a strong anti-preference for choice D.

Surprisingly, the same relative preference ordering is not only maintained when evaluated with CF+CoT, but it is magnified. The model shows a nearly exclusive preference for answering with choice ``C'' and strong aversion to choice ``A", while choice ``D'' is never once chosen as the preferred answer choice. This is not inherently problematic, given that such a distribution may be the result of imbalanced distribution in the ground truth answers. Heuristics similar to BRP also tend to be helpful and improve accuracy for some tasks \citep{wang2024not}. As we can see in the bottom row of Figure \ref{fig:dist-ridge}, however, the ground truth answer distribution for MMLU is remarkably balanced, suggesting CF+CoT has tended to decrease the models similarity to the ground truth.

\begin{figure*}[t]
    \centering
    \includegraphics[width=\linewidth]{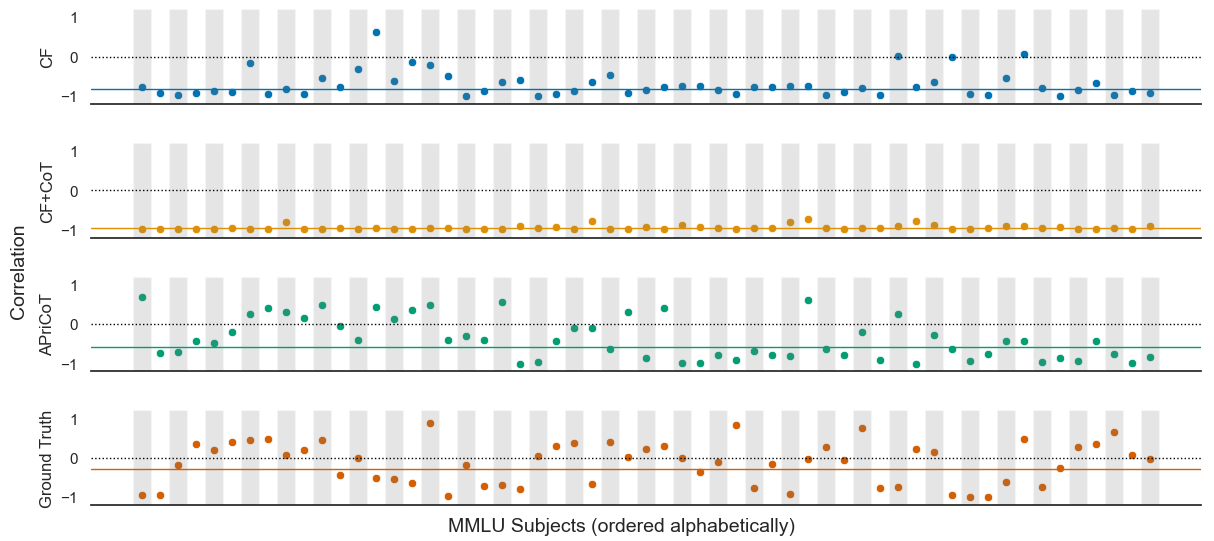}
    \caption{Pearson's $r$ correlation between average BRP per answer choice and answer choice distribution for each subject in MMLU for CF prompting with CF (blue), CF+CoT (yellow), and APriCoT (green). Lines represent the combined correlation (average Fisher's z) for each prompt method. An \textbf{ideal} language model and prompting method should have \textbf{correlation} with the BRP \textbf{consistent} with that of \textbf{ground truth} for the MMLU (red). \textbf{Best:} APriCoT has variance least explained by BRP ($r^2 = 0.35$) and is most similar to the ground truth (Wilcoxon = 521). \textbf{Worst:} CF+CoT has variance almost completely explained by BRP ($r^2 = 0.93$) and is least similar to the ground truth (Wilcoxon = 15).}
    \label{fig:dist-corr}
\end{figure*}

To further explore this, we compute the Pearson's r correlation between the chosen answer choice distribution and answer choice BRP. Correlation is computed per subject and aggregated via Fisher-z transformation before re-transformation into a summary Pearson r correlation \citep{corey1998averaging}. Results from this process are shown in Figure \ref{fig:dist-corr}. Both CF+CoT and CF answer choice distributions tend to strongly correlate with BRP ($r^2=0.68$ for CF). This is, again, especially pronounced with CF+CoT, with nearly perfect correlation across all subjects ($r^2=0.93$). This suggests that CF+CoT answers tend to be much more frequently impacted by the prior expectation of the model than even CF answers. We suspect this is a result of a confirmation bias-like effect. 

We hypothesize that the model is prone to System-1 selection of a focal point that is reinforced by subsequent CoT reasoning. This seems to be in line with previous work by \citet{turpin2024language, chua2024bias} that suggests that models tend to rationalize answers induced early in the reasoning process. Confirmation bias is apparent in the strong shifts in BRP vs answer correlation which show that CF+CoT tends to be dissimilar from the ground truth and explained by the BRP.

For completeness, we replicate the findings in \citet{moore2024base} by computing the BRP using CF prompts and perform the analysis reported for cloze BRP in Figure \ref{fig:dist-corr}. In line with that work, we find that the correlations are weak between cloze BRP and CF overt behavior ($r^2 = 0.11$ for CF, $r^2 = 0.0008$ for CoT, $r^2 = 0.12$ for APriCoT).

\subsection{Discussion: CF+CoT}

In \citet{prystawski2024think} CoT is interpreted as a method facilitating traversal of a Bayes net of related concepts. By iteratively traversing collocated concepts, facilitating connection between initially distant concepts. Essentially, CoT is a mechanism for traversing semantic space. However, this interpretation equally applies when the direction of traversal does not lead to the correct answer due to bias. 

This theoretically supports our empirical findings that CF+CoT may result in a confirmation bias. Similarly, this suggests that APriCoT is an improvement due solely to the fact that we agnostically provide directions of traversal and then require the model to provide commentary on the coherence of the paths to arrive at the best candidate answer, leading to decreased bias and increased accuracy. 

A strange observation to note is that these strong correlations between BRP and choice selection in CF+CoT are negative. This is highly counter-intuitive, as it suggests that the more preference a model has for an answer choice in the abstract, the less likely it is to be chosen in practice. We refer to this as anti-confirmation bias. We presently offer no mechanistic explanation for why this happens, but the strength of the correlations provide strong evidence for its real presence.

\section{Counterfactual APriCoT}

In order to address the shortcomings identified in the CF+CoT results, we propose an alternative method to induce reasoning in models we dub Agnostically Primed Chain-of-Thought (APriCoT). APriCoT differs from CF+CoT in two important ways (summarized in Table \ref{tab:prompts}). First, CoT is elicited in combination with one of the candidate answer choices. In effect, the model is told to evaluate the correctness of one of the answer choices. Second, each post-CoT context is used to evaluate only the same answer choice with which the CoT was primed. Each answer choice is reasoned over and queried independently for its validity. 

\subsection{Methods: APriCoT}
APriCoT is evaluated on the same MMLU question set as CF+CoT with the same 100 maximum token generation limit, 10 iterations for robustness, and answer choice selection method. All other evaluation aspects, including the maximum CoT token length, the canary completion, and the number of iterations remain unchanged from CF+CoT.


\subsection{Results: APriCoT}

Figure \ref{fig:dist-ridge} shows the total answer distribution for APriCoT. It is visibly clear that the choice distribution is well balanced and closely matches that of the ground truth answers. 

We evaluate APriCoT for its susceptibility to BRP effects. In the third row of Figure \ref{fig:dist-corr}, we see that across most subjects, the answer distribution has diminished correlation with BRP compared to CF+CoT and CF. We use the Fisher's-z transformation to aggregate correlation across subjects. The results suggest that BRP contributes $r^2 = 35\%$ of the variance across answer choices compared with $r^2 = 93\%$ for CF+CoT. APriCoT correlation is thus much closer to the correlation between ground truth and BRP ($r^2=0.08$).

We also compare BRP correlation with CF+CoT and APriCoT to the ground truth correlation with BRP. An ideal predictor would be indistinguishable from ground truth. We use the Wilcoxon signed-rank test to measure the dissimilarity between the correlations of the ground truth with CF+CoT and APriCoT correlations. The resulting test statistics are $T=15$ for CF+CoT and $T=512$ for APriCoT, suggesting that correlations for APriCoT match the ground truth much closer than for CF+CoT. This suggests that APriCoT tends to mitigate the CoT confirmation bias as it tends to be more similar to ground truth and uncorrelated with BRP.



Finally, we evaluated the effect of APriCoT on the accuracy of the model. We expected that the accuracy of the model should improve due to a combination of noise reduction from BRP mitigation and a more sophisticated deliberative mechanism. The result is an accuracy of $37.96\%$ for CF. Likely due to its susceptibility to BRP effects, CF+CoT saw a slight reduction to $35.56\%$, while APriCoT improved to $42.79\%$.

It should be noted that all results reported herein show substantially reduced accuracy on MMLU than was reported at the release of LLaMa 3.1 \citep{dubey2024llama}. This likely stems from differences in evaluation methodology. We do not have access to the specific methodology used to evaluate LLaMa 3.1. Further, previous work has shown conclusively that subtle changes such as CF prompting in lieu of cloze testing can affect model accuracy - a weakness still present in current LLMs \citep{moore2024base}. The seemingly contradictory results between our work and that of \citet{dubey2024llama} is evidence that the models and the behaviors they exhibit are complex and multifaceted. The difficulty of isolating behaviors is one of the primary motivations behind this work.

\begin{figure}[H]
    \centering
    \includegraphics[height=2in]{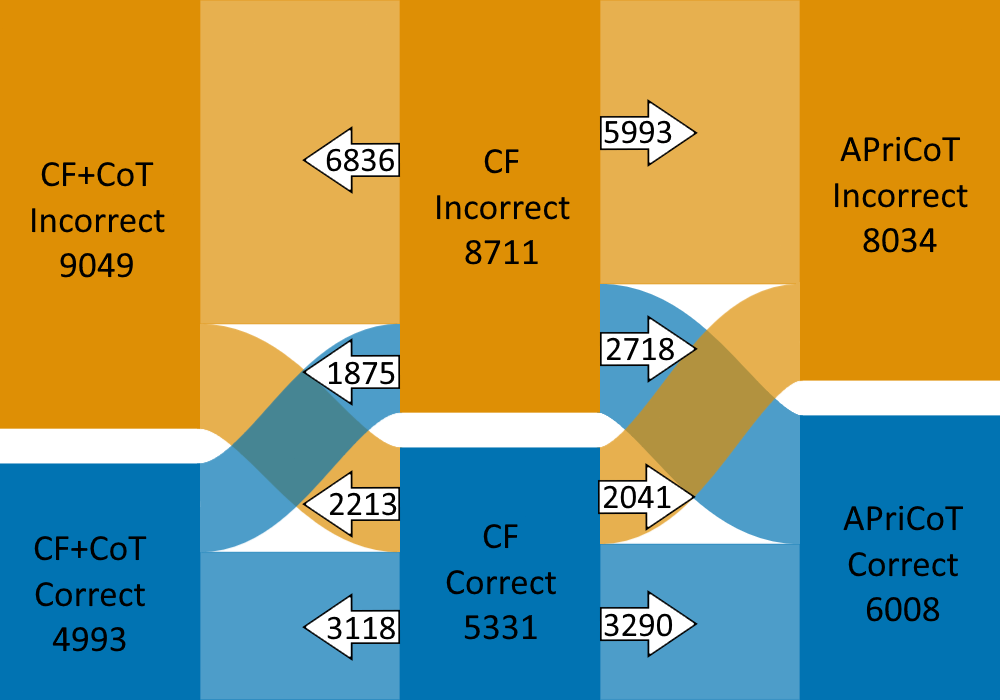}
    \caption{Flow analysis of how CF+CoT and APriCoT each changes CF performance on the MMLU.
    APriCoT has the highest accuracy at 43\% with CF+CoT = 36\% and CF = 38\%.}
    \label{fig:acc-sankey}
\end{figure}


While APriCoT improves model performance in addition to mitigating surface level biases, the results improve further with slight decomposition, as shown in Figure \ref{fig:acc-sankey}. When tracked per-question, we see that for every question that was answered correctly with CF, APriCoT is more likely to also be correct and less likely to be incorrect compared to CF+CoT. To an even greater degree, questions answered incorrectly with CF are more likely to be answered correct and less likely to be answered incorrectly using APriCoT than CF+CoT.

\section{Conclusion}
In this work, we have attempted to mitigate a simple, but pernicious, bias in LLM behavior that complicates research into other behaviors and abilities. This bias, known as base-rate bias, was shown both here and in previous work to strongly predict model behavior on the MMLU benchmark, which is a common measure of LLM factual knowledge and language understanding. We showed that induction of more complex reasoning via the commonly employed CoT prompting strategy surprisingly exacerbates the BRP bias, which we argue is the result of confirmation bias-like behavior that approximates \textit{thinking fast}.

We proposed and evaluated a novel prompting method called Agnostically Primed Chain-of-Thought, otherwise known as APriCoT. The method induces more deliberation, more closely approximating \textit{slow thinking}. Specifically, we showed that APriCoT both significantly reduces, though still does not fully eliminate, BRP effects and noticeably improves model performance on the MMLU task. We hope for this method to be a useful step toward mitigating unintended heuristics in LLM behavior research and that it might inspire useful methods in LLM-based complex reasoning systems.

\subsection{Future Work}
Future work should investigate whether APriCoT is appropriate for use in more complex reasoning tasks like search or planning. In addition, APriCoT, despite being influenced by CF prompting in its creation, makes no assumptions about the evaluation method used. As such, future work should investigate whether APriCoT yields similar improvements for cloze testing and other evaluation methods.
    
APriCoT is currently only defined for tasks with a closed set of options. Future work may investigate methods that permit open ended questions, such as dynamic option generation by the model itself. 

This work also introduces or highlights multiple theoretical questions regarding mechanistic explanations for LLM behavior that warrant further study. These include confirmation bias, the dependence of CF overt behavior on cloze BRPs, and further work evaluating the nature of LLM reasoning.

\subsection{Potential Experimental Bias}
In this work, we strove to maintain a valid, robust methodology at all stages. We recognize, however, that biases and assumptions we hold may affect the work herein and limit the generalizability of any results reported. In this section, we identify and address those biases of which we are aware to aid in the application and extension of this research.

This work is limited to experiments on a specific model and dataset, each chosen with care. The model, Meta LLaMa 3.1 8B \citep{dubey2024llama}, was the state of the art in small scale open-weight LLMs at experimentation time. We limited experiments to open-weight models to maximize reproducibility. Resource constraints limited experimentation to the smaller 8B variant. While we recognize that the results reported herein may not generalize to larger variants, we believe that they represent important findings and hope that these results inspire further study on a wider variety of models. Similarly, the MMLU benchmark is a mainstay in the evaluation of new LLMs at release. We believe this makes any results concerning MMLU's efficacy and methods for ensuring its efficacy to be of importance to the research community.

\subsection{Limitations}

APriCoT method is notably computationally expensive, increasing inference costs by a factor equal to the number of choices presented (4 in the case of MMLU). We stress that this method is intended only for model evaluation, not end-user applications. We believe that a one-time increase in cost of evaluation is justified in exchange for a more trustworthy and transparent gauge of model capabilities. We also hope that future work may provide similar benefits without the increase in computational expense.

\section{Technical Details}
All experiments were performed with $\sim$100 hours on a Google Colab NVIDIA A100 GPU. All code and data are available at \verb|www.github.com/KyleAMoore/APriCoT_Experiments|.

\bibliographystyle{apacite}

\setlength{\bibleftmargin}{.125in}
\setlength{\bibindent}{-\bibleftmargin}

\bibliography{ref}

\end{document}